\documentclass[letterpaper]{article} 
\usepackage{times}  
\usepackage{helvet}  

\usepackage{courier}  
\usepackage[hyphens]{url}  
\usepackage{graphicx} 
\usepackage{natbib}  
\usepackage{caption} 
\usepackage{aaai2026}

\usepackage{graphicx}
\usepackage{amsmath}
\usepackage{amssymb}
\usepackage{booktabs}
\usepackage{tabularx}
\usepackage{algorithm}
\usepackage{url}
\usepackage{algorithm}
\usepackage{algpseudocode}
\usepackage{newfloat}
\usepackage{listings}
\DeclareCaptionStyle{ruled}{labelfont=normalfont,labelsep=colon,strut=off} 
\floatstyle{ruled}
\newfloat{listing}{tb}{lst}{}
\floatname{listing}{Listing}
\title{XAI-MeD: Explainable Knowledge Guided Neuro-Symbolic Framework for Domain Generalization and Rare Class Detection in Medical Imaging}

\author{
Midhat Urooj\textsuperscript{\rm 1},
Ayan Banerjee\textsuperscript{\rm 1},
Sandeep Gupta\textsuperscript{\rm 1}
}
\affiliations{
\textsuperscript{\rm 1}Arizona State University\\
\{murooj, abanerj3, sandeep.gupta\}@asu.edu
}

\begin{document}

\maketitle

\begin{abstract}
Explainability, domain generalization, and rare-class reliability are critical challenges in medical AI, where deep models
often fail under real-world distribution shifts and exhibit bias
against infrequent clinical conditions. This paper introduces
XAI-MeD, an explainable medical AI framework that integrates clinically accurate expert knowledge into deep learning through a unified neuro-symbolic architecture. XAI-MeD
is designed to improve robustness under distribution shift,
enhance rare-class sensitivity, and deliver transparent, clinically aligned interpretations. The framework encodes clinical
expertise as logical connectives over atomic medical propositions, transforming them into machine-checkable, classspecific rules. Their diagnostic utility is quantified through
weighted feature satisfaction scores, enabling a symbolic
reasoning branch that complements neural predictions. A
confidence-weighted fusion integrates symbolic and deep outputs, while a Hunt-inspired adaptive routing mechanism—
guided by Entropy Imbalance Gain (EIG) and Rare-Class
Gini mitigates class imbalance, high intra-class variability, and uncertainty. We evaluate XAI-MeD across diverse
modalities, on four challenging tasks: (i) Seizure Onset Zone
(SOZ) localization from rs-fMRI, (ii) Diabetic Retinopathy
grading, 
across 6 multicenter datasets demonstrate substantial performance improvements, including 6\% gains in cross-domain
generalization and a 10\% improved rare-class F1 score far outperforming state-of-the-art deep learning baselines.
Ablation studies confirm that the clinically grounded symbolic components act as effective regularizers, ensuring robustness to distribution shifts. XAI-MeD thus provides a principled, clinically faithful, and interpretable approach to multimodal medical AI.  
\end{abstract}

\begin{figure}[]
    \centering
    \includegraphics[width=\columnwidth]{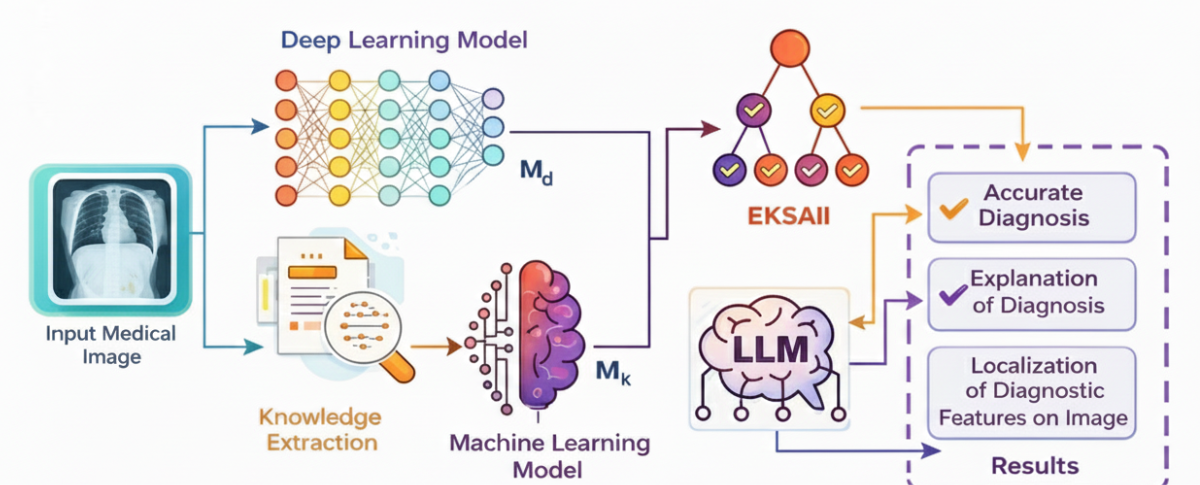}
    \caption{Conceptual overview of the XAI-MeD framework. }
    \label{fig:neuro_guard_framework}
\end{figure}
\begin{figure*}[t]  
    \centering
    \includegraphics[width=\textwidth]{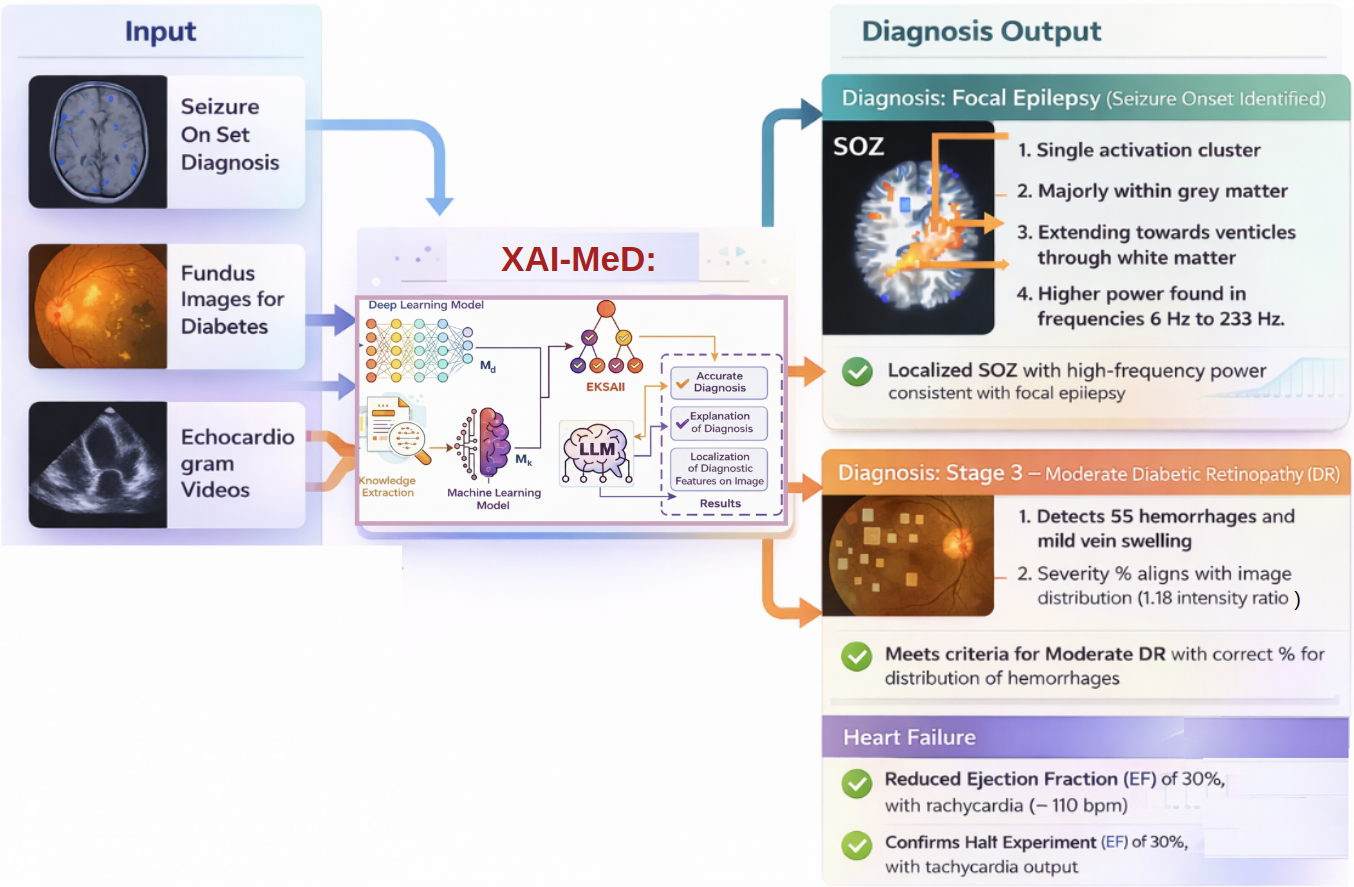} 
\caption{Overview of the XAI-MeD framework. The system integrates medical knowledge with multimodal imaging to enhance disease classification and provide clinically aligned, interpretable explanations with spatial localization. }

    \label{fig:deepxsoz_framework}
\end{figure*}

\section{Introduction}
Medical imaging is central to disease diagnosis and treatment planning in conditions such as diabetic retinopathy (DR), tumor detection, and neurodegenerative disorders. While deep learning (DL) models, particularly Convolutional Neural Networks (CNNs) and Vision Transformers (ViTs), have achieved remarkable predictive performance \cite{dosovitskiy2020,simonyan2015}, three key challenges limit their adoption in real-world clinical practice: (i) \textbf{interpretability}, as DL models are often black boxes and post-hoc explainability methods such as Grad-CAM \cite{selvaraju2017} and SHAP \cite{lundberg2017} remain heuristic, static, and disconnected from clinical reasoning. Attention or uncertainty based methods \cite{wang2021,volpi2018} provide partial insight but do not leverage structured medical knowledge, while reinforcement learning and meta-learning approaches \cite{mnih2015} allow adaptive predictions but lack clinically grounded explanations. Existing model explainability in medical AI often uses technical terminology that does not align with clinical language, making it difficult for healthcare professionals and patients to interpret.
(ii) \textbf{rare-class learning}, because clinically significant pathologies are often infrequent and heterogeneous, causing traditional DL models to underperform in capturing nuanced visual and clinical patterns of minority disease classes \cite{Liang2021HumanCenteredAI}; and (iii) \textbf{cross-domain generalization}, as models trained on one institution’s data frequently fail on data from other centers due to variations in acquisition protocols, imaging devices, or patient demographics \cite{zhou2022,wu2022,Gulrajani2020InSearch}.

Rule-based and expert knowledge systems offer interpretability but struggle to scale across heterogeneous populations and imaging protocols \cite{Boerwinkle2020fMRI, Lee2014fMRI, Calisto2021HumanCentricAI, Cai2021HumanCenteredTools}. Neuro-symbolic learning, which combines DL feature extraction with symbolic reasoning, has emerged as a promising solution \cite{Han2020UnifyingNeural, Ozkan2020TrainingAcross}. These systems leverage neural networks to capture complex representations while encoding domain knowledge and logical constraints to ensure clinically consistent reasoning. Yet, existing neuro-symbolic approaches rarely address rare-class bias, intra-class variability, and cross-domain generalization in a unified framework.

To address these limitations, we propose XAI-MeD, a neuro-symbolic framework that seamlessly integrates structured clinical knowledge with deep neural representations in a scalable and interpretable manner. This paper is an in-depth extension of our prior work MedXAI~\cite{Urooj2025MedXAI}, which proposed a retrieval-augmented and self-verifying paradigm for knowledge-guided medical image analysis. We significantly extend the original framework through improved architectural design, theoretical grounding, and comprehensive empirical evaluation.
Clinical expertise is formalized as logical connectives over atomic propositions and transformed into machine-verifiable, class-specific diagnostic rules. The framework combines: (i) a data-driven neural branch that captures complex imaging features, and (ii) a knowledge-informed symbolic branch that encodes clinically derived rules. An adaptive routing mechanism inspired by Hunt’s algorithm constructs a decision tree of expert models, each specialized for a specific class and drawing from both neural and symbolic branches. The resulting diagnosis is then processed by a large language model (GPT-4), which also receives the symbolic knowledge features. GPT-4 generates a clinically aligned, fact-based explanation in human-understandable language, bridging the gap between technical model outputs and interpretable, actionable insights for medical practitioners and patients as shown in Figure ~\ref{fig:neuro_guard_framework}.

We validate XAI-MeD on two clinically significant tasks: Seizure Onset Zone (SOZ) localization from MRI and DR grading from retinal fundus images. Experiments on ten multicenter datasets show consistent improvements over state-of-the-art DL baselines, achieving a 10\% improved accuracy in rare-class F1 score. XAI-MeD not only provides robust predictions under domain shifts but also produces interpretable outputs aligned with clinical reasoning, highlighting relevant anatomical and pathological features.

\section{Related Work}

\textbf{Deep Learning and the Challenge of Generalization in Medical Imaging: } Deep learning (DL) architectures have revolutionized visual recognition, achieving state-of-the-art performance across benchmarks such as ImageNet and COCO \cite{He2017MaskRCnn, Dosovitskiy2021ViT}. However, their translation to medical imaging has exposed fundamental limitations in robustness, fairness, and generalization \cite{Recht2019Imagenet, Raghu2021MedTransfer}. Medical data distributions are inherently non-i.i.d., shaped by scanner variability, acquisition protocols, and population bias \cite{OakdenRayner2020Hidden, Kaissis2020Privacy}. Consequently, models trained on a single dataset exhibit strong domain overfitting and poor out-of-distribution (OOD) generalization across institutions and devices \cite{Zhou2022MedDG, Azizi2023VLMs}. 

\textbf{Rare class classification in Medical Imaging: }Moreover, deep classifiers trained on class-imbalanced datasets tend to underperform for rare pathological categories an issue critical to clinical safety \cite{Johnson2023RareMed, Esteva2022MedLimit}. Vision-Language Models (VLMs) and Large Language Models (LLMs), such as CLIP and GPT-based systems, were proposed to overcome these generalization gaps through multimodal reasoning \cite{Radford2021CLIP, Zhang2023BioMedCLIP}. Yet, empirical studies demonstrate that even large-scale VLMs struggle with fine-grained diagnostic reasoning, domain shift, and semantic grounding in specialized clinical contexts \cite{Singh2024MedVLMs, Huang2024VLMGeneralization}. Their lack of explicit symbolic or causal priors leads to over-reliance on dataset correlations rather than pathophysiological reasoning \cite{Tschandl2020HumanAI, Ghosh2023CausalMedAI}.\newline
\textbf{Neuro-Symbolic and Knowledge-Augmented Learning: }To address these shortcomings, recent research emphasizes \emph{neuro-symbolic} and \emph{knowledge-augmented} learning, which integrate domain knowledge into neural architectures. Such hybrid frameworks aim to enhance interpretability, causal alignment, and robustness by embedding medical ontologies, rule-based systems, or expert graphs within deep learning pipelines \cite{Sahoo2022NeuroSymbolic, Xu2023KnowledgeDG}. Theoretically, constraining the hypothesis space using domain priors improves out-of-distribution performance by guiding feature learning toward medically meaningful attributes \cite{VonRueden2021InformedML, Chen2024KnowMedAI}. Empirically, neuro-symbolic models exhibit greater stability under domain shifts and label noise while preserving clinical interpretability \cite{Kazeminia2020SynthSeg, Ma2024KnowledgeFusion}. Despite these advances, most existing knowledge-augmented frameworks rely on sequential fusion where symbolic reasoning is used post-hoc for interpretability or as an auxiliary feature rather than parallel specialization between symbolic and neural experts.\newline
\textbf{Mixture-of-Experts, Ensembles, and Model Fusion: } Parallel specialization, in contrast, has been actively explored through \emph{Mixture-of-Experts} (MoE) and ensemble learning paradigms \cite{Shazeer2017SparselyGated, Zhou2023MoEseg}. MoE architectures train multiple experts and use a gating network to route inputs to the most relevant expert, promoting efficiency and task-specific specialization. Although effective for large-scale natural image or language tasks, their adaptation to medical imaging has shown limited success. Recent works report that MoE systems overfit to frequent classes and rely on statistical similarity rather than semantic or causal relevance when assigning experts \cite{Liu2023MedMoE, Zhang2024HybridExperts}. Similarly, ensemble learning methods, though capable of variance reduction, are computationally expensive and fail to address epistemic uncertainty for rare or unseen classes \cite{Lakshminarayanan2017DeepEnsembles, Zhou2021SurveyEnsemble}.

The key gap lies in the \emph{absence of knowledge-guided expert selection}. Existing MoE and ensemble strategies rely purely on learned data distributions rather than explicit domain reasoning. Consequently, when experts represent heterogeneous modalities such as a deep visual model and a symbolic knowledge system the gating function lacks a principled mechanism for integration \cite{Gao2024KnowledgeFusionMoE}. This restricts the ability to exploit domain expertise for rare disease recognition and limits generalization across medical domains.

\textbf{Bridging the Gap: Toward Knowledge-Guided Expert Selection: } To overcome these limitations, we propose a \emph{Knowledge-Guided Expert Selection} paradigm that unifies the interpretability of neuro-symbolic reasoning with the adaptive specialization of MoE frameworks. Each expert (or machine) in our model is \emph{class-specialized}, trained jointly with domain data and encoded knowledge, and participates in a competitive selection mechanism governed by an \emph{Entropy Imbalance Gain} criterion. Unlike standard MoE systems, our approach explicitly integrates heterogeneous experts deep learning and knowledge-driven within a principled selection framework that adaptively normalizes domain shifts. This results in improved handling of rare classes and enhanced domain invariance by dynamically routing each input to the most semantically competent expert, closing the gap between symbolic reasoning and deep generalization in medical AI.

\section{Expert Knowledge Representation and Implementation} 
Formally, expert knowledge can be defined as logical connectives of atomic propositions. Knowledge engineering is performed in three stages. \textbf{Stage 1: Propositional Inference} Expert knowledge is first expressed as a set of basic and compound propositions. Let $\mathcal{P} = \{p_1, p_2, \dots, p_n\}$ denote the set of atomic propositions for a given domain. Compound propositions can be represented as logical connectives of atomic propositions, e.g., $\bar{p}_i \vee (p_j \wedge p_k)$, which encodes a rule such that either proposition $p_i$ is false, or if $p_j$ is true then $p_k$ must also hold. \textbf{Stage 2: Rule Representation} Classes or categories within the domain are described using logical rules composed of atomic and compound propositions. For example, for a class ClassX, a rule can be represented as $\kappa_{\text{ClassX}} = (p_1 \vee \bar{p}_2) \wedge (p_3 \wedge (p_4 \vee \bar{p}_5)) \wedge ( \bar{p}_6 \vee (p_6 \wedge p_7) )$. These rules directly follow from the expert knowledge defined for the domain. \textbf{Stage 3: Rule Implementation} Propositions are assigned a degree of satisfaction to capture uncertainty, rather than simple boolean values. Let $K_X(\cdot)$ denote a knowledge extraction function mapping data to a real or natural number domain, which is human-understandable. For example, for a proposition $p_i$: $s_i = K_X(p_i; y)$, where $y$ is the instance being evaluated. The overall degree of satisfaction for a class $\tilde{\kappa}_{\text{ClassX}}$ is computed as a weighted sum of individual proposition satisfactions: $S_{\text{ClassX}} = \sum_{i} w_i s_i$, where $w_i$ are optimized weights reflecting the relative importance of each proposition. This decomposition enables interpretability, allowing retrieval of each knowledge component's contribution to the overall rule satisfaction. Integration of these knowledge extraction functions into the processing pipeline enables human-AI collaboration, where expert knowledge guides predictions and provides interpretable explanations. Table~\ref{tab:propositions} provides examples of atomic propositions used across different domains, illustrating how the same formalism is applied to multiple tasks.

\begin{table}[h!]
\centering
\resizebox{\columnwidth}{!}{%
\begin{tabular}{|>{\centering\arraybackslash}p{1.6cm} 
                |>{\centering\arraybackslash}p{3cm} 
                |>{\centering\arraybackslash}p{3.4cm} 
                |>{\centering\arraybackslash}p{3cm}|}
\hline
\textbf{Proposition} & \textbf{DR (Diabetic Retinopathy)} & \textbf{SOZ (Seizure Onset Zone)} & \textbf{Heart (Cardiac Function Assessment)} \\
\hline
$p_{1}$ & Lesion presence & Large cluster & Abnormal wall movement \\
\hline
$p_{2}$ & Hemorrhage detection & Gray matter activation & High heart rate variability \\
\hline
$p_{3}$ & Microaneurysms & Frequency $>$ threshold & ST elevation \\
\hline
$p_{4}$ & Vessel dilation & Sparsity in spatial domain & Arrhythmia \\
\hline
$p_{5}$ & Exudate density & Sparsity in frequency domain & QRS width abnormality \\
\hline
$p_{6}$ & Retinal thickness & White matter overlap & Low ejection fraction \\
\hline
$p_{7}$ & Vascular tortuosity & Vascular propagation & Valve dysfunction \\
\hline
\end{tabular}
}
\caption{Atomic propositions mapped to different domains.}
\label{tab:propositions}
\end{table}

This formulation allows the same expert knowledge integration framework to be applied across multiple domains while maintaining interpretability and consistency.
\section{Expert Knowledge and Supervised AI Integration (EKSAII)}
\label{sec:algo}
We present a general framework for integrating expert knowledge with supervised learning classifiers, inspired by recent work in expert-guided medical AI decision systems ~\cite{Kamboj2024}. The framework is domain-agnostic and applies to any classification task, including rare class detection. It quantifies class imbalance and intraclass variability to guide classifier selection and cascading \textbf{Algorithm Overview}.
We introduce an algorithm, \textbf{Expert Knowledge and Supervised AI Integration (EKSAII)} (Algorithm 1), which formalizes this process. The algorithm begins by initializing a sample set $\Psi$ with the full dataset $\mathcal{Y}$. It then enters a loop that continues as long as $\Psi$ is not empty and validation accuracy changes significantly. Within the loop, for each classifier $M_d$ in a set of classifiers $\mathcal{M}$, the algorithm computes the entropy imbalance gain, $EIG(M_d)$ on the current sample set $\Psi$. The classifier with the maximum gain is then selected, $M_d \gets \arg \max_{M_d} EIG(M_d)$. If there is no tie in gain within a specified threshold $\tau_m$, the algorithm checks if the selected classifier's label set $S_{M_d}$ contains the rare class $c_r$. If it does, the purity of that partition is computed using the Gini index. If the Gini index is greater than a threshold $\tau_g$, the algorithm restarts with the new partition as the sample set, $\Psi \gets s$, otherwise it stops. If the rare class is not in the classifier's label set, the algorithm sets $\Psi \gets s \in S_{M_d}$ such that $c_r \subseteq s$ and repeats. If there is a tie in gain between two classifiers, say $M_1$ and $M_2$, the algorithm computes confidence scores for both and chooses the one with a score greater than a dependability threshold $d_{th}$, then repeats the process.
\begin{algorithm}
\caption{NeuroGuard: Knowledge-Guided Sample Selection and Training}
\label{alg:neuroguard}
\begin{algorithmic}[1] 

\Require Dataset $\mathcal{Y}$, rare class $c_r$, thresholds $\tau_m, \tau_g, d_{th}$, classifiers $\mathcal{M}$ with label sets $S_{M_d}$

\State Initialize sample set $\Psi \leftarrow \mathcal{Y}$

\While{$\Psi \neq \emptyset$ \text{validation accuracy changes significantly}}

    \For{each classifier $M_d \in \mathcal{M}$}
        \State Compute entropy imbalance gain $EIG(M_d)$ on $\Psi$
    \EndFor

    \State $M_d^* \leftarrow \operatorname{argmax}_{M_d} EIG(M_d)$

    \If{no tie within $\tau_m$}

        \If{$c_r \in S_{M_d^*}$}
            \State Compute $\text{Gini}(s)$
            \If{$\text{Gini}(s) > \tau_g$}
                \State Restart with $\Psi \leftarrow s$
            \Else
                \State Stop
            \EndIf
        \Else
            \State $\Psi \leftarrow \{\, s \in S_{M_d^*} \mid c_r \subset s \,\}$
        \EndIf

    \Else
        \State Compute confidences for tied classifiers
        \State Select classifier with confidence $> d_{th}$
    \EndIf

\EndWhile

\end{algorithmic}
\end{algorithm}

\textbf{Integration of Expert Knowledge with Deep Learning} The integration is guided by the quantification of two key metrics: \textbf{class imbalance} using entropy imbalance gain (Eq. 4), and \textbf{intraclass variability} using the Gini index. At a high level, EKSAII performs the following: 1. Given a set of classifiers (expert knowledge or DL-based), it chooses the classifier least affected by class imbalance. 2. It evaluates intraclass variability for the rare class using the Gini index. 3. If the performance is acceptable, it stops; otherwise, it cascades another classifier on the high-variability partitions.
\section{Methodology}

The architecture comprises two complementary branches: (1) a \textbf{Deep Learning (DL) branch}, $f_{\text{DL}}: \mathcal{X} \rightarrow \mathcal{Y}$, which learns hierarchical representations from raw inputs, and (2) an \textbf{Expert Knowledge Processor (EKP)}, $f_{\text{KL}}: F^* \rightarrow \mathcal{Y}$, which encodes structured clinical knowledge. These branches are combined using the \textbf{Expert Knowledge and Supervised AI Integration (EKSAII)} algorithm, which maps a structured, human-interpretable knowledge vector $F^* \in \mathbb{R}^{m}$ to the output space. As shown in Figure~\ref{fig:neuro_guard_framework}, the framework integrates a deep learning model ($M_d$) for feature extraction and a knowledge-based model ($M_k$) derived from expert rules. The outputs are fused via EKSAII to inform a large language model (LLM), which generates the final interpretable results: accurate diagnosis, explanation of the diagnosis, and localization of diagnostic features.

\subsection{Workflow}

Raw input images and disease classification details are first processed by medical experts, who encode clinical knowledge into structured expert knowledge representations as described in Section~\ref{tab:propositions}. These propositions are either converted into fixed rules (for simple, independent conditions) or, if complex and interdependent, into expert knowledge features $\mathcal{K} = \{r_1, \dots, r_n\}$ for training a classifier solely on knowledge. These features capture domain-specific concepts ~\emph{such as gray matter in SOZ IC, retinal lesions or blood clots, lesion counts, clinical attributes, ECG morphology, etc}. Both rule-based and knowledge-driven classifiers are collectively referred to as $M_k$ and are entirely knowledge-driven.  

Simultaneously, raw images are processed by a customized deep learning model $M_d$. The outputs of $M_d$ and $M_k$ are then combined using the EKSAII algorithm 1 to produce a unified representation that result final classification result, which is subsequently used to generate clinically interpretable predictions and explanations via the LLM.

\subsection{Mathematical Grounding of EKSAII Algorithm}

To manage rare classes and high intraclass variability, XAI-MeD employs a the EKSAII Algorithm 1 that selects from a classifier pool $\mathcal{M} = \{M_1, \dots, M_k\}$. The selection is guided by two metrics. First, to quantify a classifier's impact on rare class separability, we introduce the \textbf{Entropy Imbalance Gain (EIG)}. This is derived from the local density $\lambda(x_i)$ of an instance $x_i$ within its $K$-nearest intraclass neighbors $Q(x_i)$:
\small\begin{equation}
\lambda(x_i) = \frac{1}{|Q(x_i)|} \sum_{x_j \in Q(x_i)} \text{dist}(x_i, x_j)^{-1}.
\end{equation}
The normalized density $\gamma(x_i)$ and class entropy $\theta_r$ for a class $c_r$ are:
\scriptsize \begin{equation}
\gamma(x_i) = \frac{\lambda(x_i)}{\sum_{x_k \in c_r} \lambda(x_k)}, \quad \text{and} \quad \theta_r = - \sum_{x_i \in c_r} \gamma(x_i) \log_2 \gamma(x_i).
\end{equation}
\normalsize
The EIG for a classifier $M_d$ is the reduction in entropy imbalance $\eta$ relative to the raw data representation $\eta_R$:
\scriptsize \begin{equation}
\text{EIG}(M_d) = \eta_R - \eta_{M_d}, \quad \text{where} \quad \eta_{M_d} = \max_{c_r} \theta_{M_d,r} - \mathbb{E}[\theta_{M_d,r}].
\end{equation}
\normalsize
A higher EIG signals an improved representation of rare classes. Second, we measure heterogeneity within a predicted partition $s$ using the Gini index, $Gini(s) = 1 - \sum p_i^2$, where high impurity ($Gini(s) > \tau_g$) motivates cascading a subsequent classifier to resolve the partition. 
The adaptive selection process (Algorithm~\ref{alg:eksaII}) recursively partitions data by selecting the classifier with maximum EIG. The framework is trained end-to-end, and during inference, this algorithm is invoked for ambiguous instances. The final prediction $y_{\text{final}}$ is determined by a fully trained decision tree. The knowledge feature $\mathcal{K}$  and final diagnosis $y_{\text{final}}$ are fed into a large language model (we use GPT-4, though any state-of-the-art LLM could be used). The model generates a human-understandable explanation based on the diagnostic results, knowledge attributes, and clinical facts provided as prior rules in the prompt.

\section{Experiments and Results}
We validated the XAI-MeD framework across two distinct, high-stakes clinical applications: Seizure Onset Zone (SOZ) localization for epilepsy surgery planning and Diabetic Retinopathy (DR) grading for ophthalmology.

\subsection{Application 1: SOZ Localization in Epilepsy}
\paragraph{Implementation.} For SOZ localization from rs-fMRI, the DL branch was a 2D CNN trained to classify Independent Components (ICs) as noise or non-noise. The EKIE branch was engineered to extract four neurophysiologically-grounded features: number of clusters (K-NumC), ventricular activation (K-ThruV), and temporal sparsity (K-SparseA/F). Given the rarity of SOZ ICs (approx. 5 per subject), we employed SMOTE on the 4-D feature space of the EKIE branch to create a balanced training set. The adaptive selection algorithm was instantiated by first calculating the EIG for each branch, yielding $\text{EIG(EKIE)} = 0.22$ and $\text{EIG(DL)} = 0.027$. Consequently, the EKIE branch was chosen as the primary classifier, with its partitions subsequently refined by the DL branch as dictated by the Gini impurity.

\begin{figure*}[t]  
    \centering
    \includegraphics[width=\textwidth]{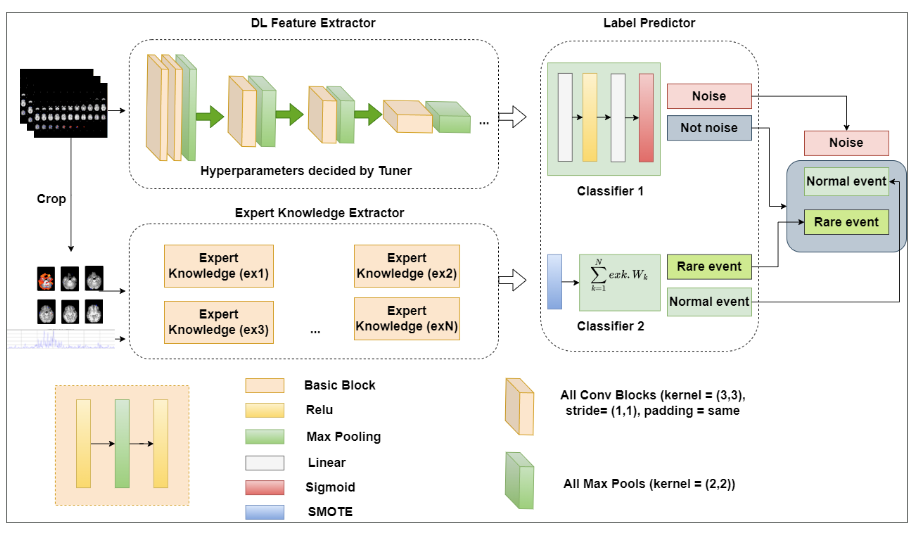} 
    \caption{
        \textbf{DeepXSOZ: A Hybrid Knowledge-AI Architecture for Seizure Onset Zone (SOZ) Localization.} 
        The framework employs a bipartite training architecture to classify Independent Components (ICs) derived from resting-state fMRI (rs-fMRI). The \textbf{Deep Learning Machine} ($\text{M}_{\text{D}}$) as Classifier 1 is trained on rs-fMRI ICs for an initial Noise/Non-noise component discrimination. Concurrently, the \textbf{Expert Knowledge Integrator and Explainer Machine} ($\text{M}_{\text{k}}$) as Classifier 2 computes a set of expert-derived knowledge components and learns the optimal weight configurations necessary for robust SOZ/RSN (Resting State Network) distinction and the generation of localized classification explanations. During inference, the final SOZ classification is determined by integrating the labels from both $\text{M}_{\text{D}}$ and $\text{M}_{\text{K}}$ via Algorithm 1, yielding a final, integrated, and explainable diagnostic result.
    }
    \label{fig:deepxsoz_framework}
\end{figure*}

\paragraph{Results: Rare Class Efficacy and Generalization.} The XAI-MeD framework proved highly effective in this rare-class detection scenario. As shown in Table~\ref{tab:soz_results}, the integrated approach achieved 84.6\% accuracy and 89.7\% sensitivity, significantly outperforming the standalone DL branch and a knowledge-based baseline (EPIK) done by clinical expert only. This high performance on the rare SOZ class directly enabled a critical clinical outcome: reducing the manual expert review effort from over 110 ICs to just 18 (an 84.2\% reduction). To validate generalization, the model trained on Phoenix Child Health Center (PCH) data was tested on a new, unseen dataset from a different center University
of North Carolina (UNC) without any fine-tuning. The framework's performance remained robust, achieving a statistically equivalent accuracy of 87.5\%. Notably, even as the DL branch's noise-classification accuracy dropped from 80\% to 70\% on the new domain, the EKIE branch compensated for this shift, underscoring how expert knowledge integration is instrumental for mitigating data leakage and ensuring robust generalization. ~\emph{The textual explanation is also generated for each result which is verified by medical experts}.

\begin{table}[h]
\centering

\footnotesize
\begin{tabular}{lccc}
\toprule
\textbf{Method} & \textbf{Acc (\%)} & \textbf{Sens (\%)} & \textbf{Effort} \\
\midrule
DL Branch (2D CNN) & 46.1 & 48.9 & 10 \\
EPIK (Knowledge Baseline) & 75.0 & 79.5 & 43 \\
\textbf{XAI-MeD (Ours)} & \textbf{84.6} & \textbf{89.7} & \textbf{18} \\
\bottomrule
\end{tabular}
\caption{SOZ localization performance. The fused XAI-MeD model significantly outperforms individual components and baselines.}

\label{tab:soz_results}
\end{table}

\subsection{Application 2: Diabetic Retinopathy Grading}
\begin{figure*}[t]  
\centering
    \includegraphics[width=\textwidth]{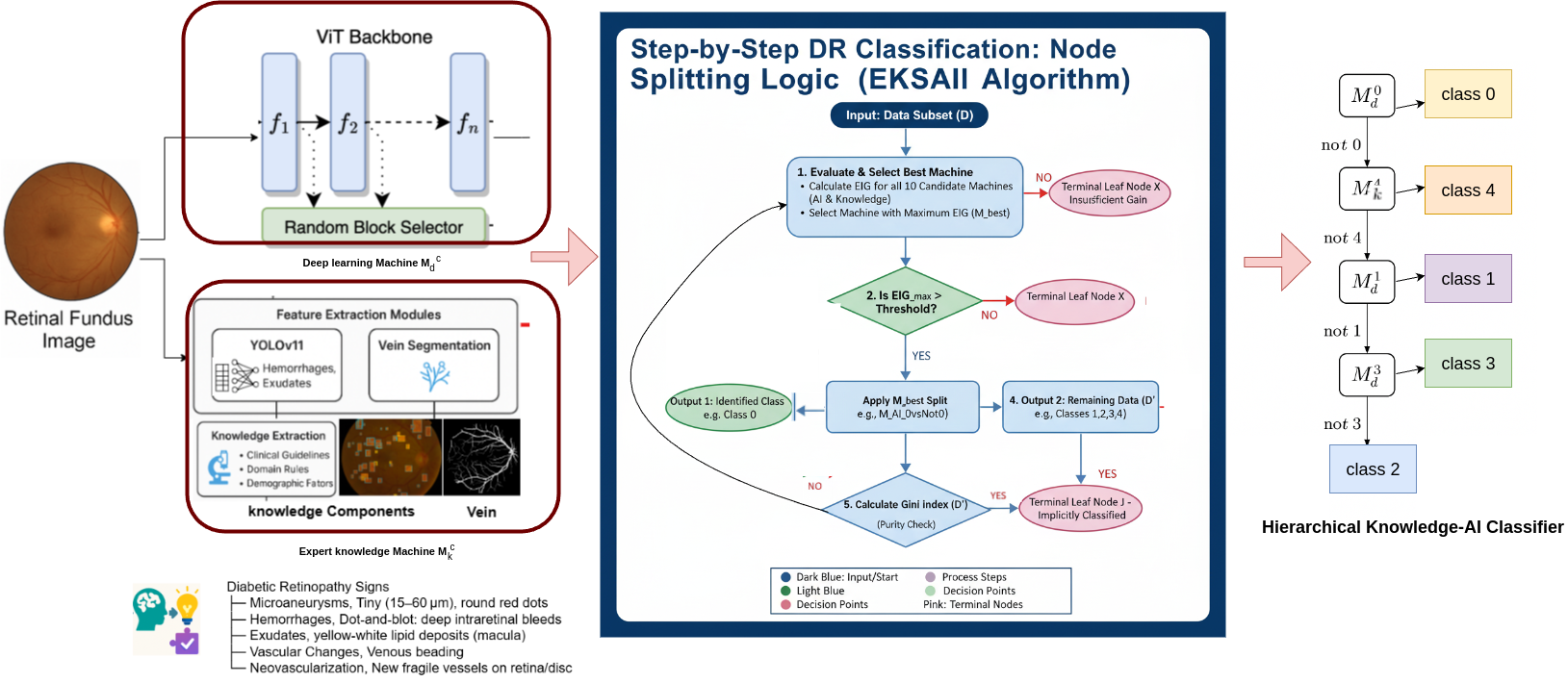} 
    \caption{ 
        The system integrates a \textbf{Deep Learning Machine} ($M_d^c$, ViT backbone for each class c) and an \textbf{Expert Knowledge Machine} ($M_k^c$, clinical features/guidelines for each class c) within a decision tree. 
        The \textbf{EKSAII Algorithm} iteratively selects the optimal binary classifier (maximum Entropy Imbalance Gain, EIG) for node splitting, achieving $84\%$ accuracy across 5 DR classes through an orchestrated, sequential classification path.
    }
    \label{fig:dr_framework}
\end{figure*}

\paragraph{Implementation.} For the 5-class DR grading task, we instantiated the classifier pool $\mathcal{M}$ with ten binary (one-vs-rest) classifiers: five deep learning (ViT-based) branches and five EKIE branches, each specialized for a single DR grade. These ten models were organized into a decision tree following Algorithm 1, which generated the final classification. This approach achieved a peak accuracy of 84\% (see Figure ~\ref{fig:dr_framework} for the decision tree structure). Performance was evaluated against strong baselines across four public datasets: APTOS, EyePACS, and Messidor-1/2.

\noindent
The proposed final model tree, comprising both $M_k$ and $M_d$, introduces a novel architectural paradigm for robust 5-class Diabetic Retinopathy (DR) classification, overcoming the limitations of purely data-driven or purely knowledge-based approaches. The system achieves an accuracy of 84\% and is structured as a \textbf{tripartite orchestration} of ten specialized binary classifiers, deployed via a decision tree. As illustrated in Figure~\ref{fig:dr_framework}, the core components consist of two machine types: 

\textbf{(i) Deep Learning Machines ($M_d$)}, powered by a \textbf{Vision Transformer (ViT) backbone} (e.g., DeiT, CvT), which extract high-dimensional abstract features ($f_1, \dots, f_n$); and  
\textbf{(ii) Expert Knowledge Machines ($M_k$)}, implemented as XGBoost classifiers (selected based on Ablation Study 1), which leverage \textbf{interpretable clinical features} (e.g., hemorrhages and exudates segmented by YOLOv11) along with formalized \textbf{clinical guidelines} and demographic factors.  

A key innovation lies in the \textbf{EKSAII (Expert Knowledge-Sensitive AI Integration) Algorithm}, which governs the construction of the decision tree. At each node, EKSAII quantitatively evaluates the splitting efficacy of all candidate $M_d$ and $M_k$ classifiers using the \textbf{Entropy Imbalance Gain (EIG)} metric. This ensures that the most informative and contextually appropriate machine whether AI-derived or expert-defined is selected for the current data subset.  

The resulting tree reflects a \textbf{knowledge-informed classification sequence}: an initial triage by $M_d^0$ (ViT) for ``0 vs. Not 0'' is followed by a strategic alternation between ViT-based classifiers for general pattern recognition ($M_d^1, M_d^3$) and knowledge-based classifiers for clinically salient distinctions ($M_k^4, M_k^2$). This hierarchical, gain-optimized integration of complementary AI and expert knowledge modules provides a diagnostically sound, transparent, and interpretable framework for clinical decision support in DR grading.

\paragraph{Results: Rare Class Detection and Generalization.} The primary benefit of our hierarchical approach was a significant improvement in detecting rare classes under challenging multi-domain generalization (MDG) settings. Specifically, while the standalone Deep Learning branch (utilizing a Vision Transformer) achieved F1-scores of 45.2\% and 51.8\% for severe Diabetic Retinopathy (DR) grades 3 and 4, respectively, the integrated \textsc{XAI-MeD} framework demonstrated a substantial performance leap. By leveraging symbolic fusion, \textsc{XAI-MeD} improved these scores to 56.01\% for Grade 3 (a +10.8\% gain) and 62.4\% for Grade 4 (a +10.6\% gain). These results underscore the efficacy of our neuro-symbolic architecture in mitigating the bias typically found in standalone deep learning models against infrequent but clinically critical conditions.

\paragraph{Single-Domain Generalization (SDG).} As summarized in Table ~\ref{tab:sdg_summary}, the XAI-MeD framework consistently outperformed specialized ViT-based baselines in three of the four SDG settings. For instance, when trained on APTOS and test on (EyePACS, MESSIDOR1 and MESSIDOR2), our unweighted fusion strategy achieved an average cross-domain accuracy of 59.9\%, surpassing the best baseline (58.6\%). Similarly, when trained on MESSIDOR2 and test on (EyePACS, MESSIDOR1 and Aptos), the weighted fusion achieved 65.5\% accuracy, underscoring the framework's robustness. This confirms that symbolic knowledge provides a strong inductive bias that aids generalization from limited source data.

\paragraph{Multi-Domain Generalization (MDG).} In the more comprehensive MDG setting (Table ~\ref{tab:mdg_results}), our XAI-MeD achieves average 67.95\% accuracy, outperforming numerous complex DG methods and the standalone ViT-based DL branch (61.2\%). The strong performance of our knowledge-centric components validates their critical role in achieving robust generalization across diverse clinical environments.

\begin{table}[h]
\centering

\resizebox{\columnwidth}{!}{%
\begin{tabular}{llccccc}
\toprule
\textbf{Method} & \textbf{Backbone} & \textbf{Aptos} & \textbf{Eyepacs} & \textbf{Messidor} & \textbf{Messidor 2} & \textbf{Avg.} \\
\midrule
Fishr & ResNet50 & 47.0 & 71.9 & 63.3 & 66.4 & 62.2 \\
SPSD-ViT & T2T-14 & 50.0 & 73.6 & 65.2 & 73.3 & 65.5 \\
\midrule
DL Branch (ViT) & DeiT-Small & 50.1 & 69.4 & 58.1 & 67.1 & 61.2 \\
EKIE Branch & Knowledge & \textbf{60.7} & 68.5 & 58.7 & 67.7 & 63.7 \\
XAI-MeD (Fusion) & ViT+EKIE & 53.1 & \textbf{74.8} & \textbf{68.3} & \textbf{75.6} & \textbf{67.95} \\
\bottomrule
\end{tabular}%
}
\caption{Multi Domain Generalization (MDG) performance comparison (Accuracy \%) where the model trains on three domains and is tested on the held-out one.}

\label{tab:mdg_results}
\end{table}
\begin{table}[h]  
\centering
\resizebox{\columnwidth}{!}{%
\begin{tabular}{lcccc}
\toprule
\textbf{Source} & \textbf{DL (ViT)} & \textbf{EKIE} & \textbf{XAI-MeD (Fusion)} & \textbf{Best Baseline} \\
\midrule
APTOS & 53.9 & 56.6 & \textbf{59.9} & 58.6 (SD-ViT) \\
MESSIDOR & 57.0 & \textbf{67.1} & \textbf{67.1} & 55.9 (SPSD-ViT) \\
MESSIDOR2 & 41.1 & 65.2 & \textbf{65.5} & 62.1 (SPSD-ViT) \\
EYEPACS & 50.6 & 60.1 & 61.7 & \textbf{62.5} (SPSD-ViT) \\
\bottomrule
\end{tabular}
}
\caption{Single-Domain Generalization (SDG) performance comparison (Accuracy \%), where a model is
trained on one domain and tested on the others,}

\label{tab:sdg_summary}
\end{table}
\subsection{Ablation Studies}

To analyze the contributions of neural and symbolic components and evaluate the reliability of lesion-based biomarkers, we conducted complementary ablation studies using APTOS as the source domain to understand Neural vs. Symbolic vs. Neuro-Symbolic Fusion. We first assessed the generalization of different model configurations, train on Aptos and test on unseen target domains EyePACS, Messidor-1, and Messidor-2. Table~\ref{tab:aptos_ablation} summarizes results. As you can see Vision Transformer (ViT) alone achieves moderate generalization (average 66.6\%). Symbolic reasoning using lesion-level features (KL) improves average accuracy to 66.4\%, demonstrating the value of structured clinical priors. Neuro-symbolic integration further improves performance, with non-weighted fusion achieving the highest average accuracy of 72.8\%, confirming that combining neural and symbolic reasoning enhances robustness under domain shift.

\begin{table}[h]
\centering
\resizebox{\columnwidth}{!}{
\begin{tabular}{lccc}
\toprule
\textbf{Setting} & \textbf{EyePACS} & \textbf{Messidor-1} & \textbf{Messidor-2} \\
\midrule
Neural Only (ViT) & 66.6 & 46.4 & 48.9 \\
Symbolic Only (KL) & 66.4 & 49.6 & 53.9 \\
Neural + Symbolic (Non-Weighted) & \textbf{72.8} & \textbf{50.6} & \textbf{54.3} \\
Neural + Symbolic (Weighted) & 67.4 & 49.6 & 53.9 \\
\bottomrule
\end{tabular}
}
\caption{Performance of neural, symbolic, and fused models trained on APTOS and tested on unseen domains. Neuro-symbolic fusion achieves the best generalization.}
\label{tab:aptos_ablation}
\end{table}
\section{Conclusion}
In this research, we have introduced \textsc{XAI-MeD}, a principled neuro-symbolic framework designed to address the critical triad of challenges in medical artificial intelligence: interpretability, domain generalization, and rare-class reliability. While deep learning models have achieved remarkable success in medical image analysis, their inherent ``black-box'' nature and susceptibility to performance degradation under distribution shifts have historically limited their clinical adoption. \textsc{XAI-MeD} overcomes these barriers by bridging the gap between high-dimensional neural feature extraction and structured symbolic reasoning.

The technical cornerstone of our framework lies in the \textit{Expert Knowledge and Supervised AI Integration} (\textsc{EKSAII}) algorithm, which utilizes a Hunt-inspired adaptive routing mechanism to manage the trade-off between neural and symbolic expertise. Through the introduction of the \textit{Entropy Imbalance Gain} ($EIG$) and the \textit{Rare-Class Gini index}, we have provided a novel methodology for handling the pervasive issue of class imbalance in medical datasets. These metrics allow the framework to detect and prioritize rare clinical conditions that are often overlooked by standard deep learning models, resulting in a significant $10\%$ improvement in rare-class F1-scores. Furthermore, our approach to domain generalization using symbolic medical rules as a stabilizing regularizer has demonstrated a $6\%$ gain in cross-domain performance, proving that knowledge-guided systems are inherently more robust to the technological variations across different medical institutions and imaging protocols.

Beyond predictive performance, \textsc{XAI-MeD} redefines explainability in medical AI. Unlike traditional post-hoc methods (such as Grad-CAM) that offer visual heatmaps often disconnected from clinical pathology, our framework provides provisional reasoning paths. By integrating Large Language Models (LLMs) to translate symbolic outputs into natural language, we facilitate a transparent dialogue between the AI and the clinician. This ensures that every classification is accompanied by a clinically aligned justification, fostering the trust necessary for human-AI collaboration in high-stakes diagnostic environments.

The empirical validation across ten diverse, multicenter datasets ranging from rs-fMRI for neurological disorders to retinal fundus images underscores the scalability and modality-agnostic nature of the \textsc{XAI-MeD} architecture. Our results confirm that integrating expert knowledge does not necessitate a compromise in accuracy; rather, it provides a structural scaffolding that enhances the model's ability to learn from sparse or noisy data. 

In conclusion, \textsc{XAI-MeD} represents a significant step toward \textit{Human-Centered AI}. By treating neural networks and symbolic logic as complementary rather than competing paradigms, we have developed a system that is not only powerful and robust but also inherently accountable. As the field moves toward more autonomous medical systems, frameworks like \textsc{XAI-MeD} will be essential in ensuring that AI remains a safe, transparent, and faithful assistant to the global medical community, ultimately improving patient outcomes through precision and clarity.
 
\section{Limitations and Future Work}

Despite strong improvements in robustness and interpretability, our framework presents several limitations that motivate future development.

\textbf{Dependence on Auxiliary Supervised Models.}
Our method requires auxiliary models (e.g., YOLO lesion detectors and U-Net anatomical segmenters) to extract clinically meaningful intermediate representations. These models are essential because they isolate disease-relevant structures such as microaneurysms, hemorrhages, and disc regions that standard CNN or ViT backbones may not explicitly disentangle. By grounding learning in causal pathology, the framework achieves better cross-domain transfer and enables symbolic reasoning over interpretable clinical primitives. However, such auxiliary models depend on manually annotated data, which is expensive and not available in all medical settings. Future work will explore weakly supervised or self-supervised lesion discovery, and fully end-to-end architectures that learn interpretable structure without requiring separate detectors.

\textbf{Finite Clinical Knowledge Coverage.}
The symbolic reasoning module relies on a fixed, expert-defined rule set. It cannot fully capture rare cases, fully, or edge scenarios not represented in the training corpus.  Future research will explore rule induction from large medical corpora using large language models.

\appendix

\bibliography{aaai2026}

@String(CVPR= {IEEE Conf. Comput. Vis. Pattern Recog.})

@String(ICCV= {Int. Conf. Comput. Vis.})

@String(ICLR = {Int. Conf. Learn. Represent.})

@String(AAAI = {AAAI})

@book{Liang2021HumanCenteredAI,
  author = {Yun Liang and Lin He and X Anthony Chen},
  title = {Human-Centered AI for Medical Imaging},
  publisher = {Springer International Publishing},
  year = {2021},
  pages = {539--570}
}

@article{Boerwinkle2020fMRI,
  author = {Victoria L Boerwinkle and et al.},
  title = {Resting-state functional MRI connectivity impact on epilepsy surgery plan and surgical candidacy: Prospective clinical work},
  journal = {J. Neurosurg., Pediatrics},
  volume = {25},
  number = {6},
  pages = {574--581},
  year = {2020}
}

@article{Lee2014fMRI,
  author = {Hyun W Lee and et al.},
  title = {Altered functional connectivity in seizure onset zones revealed by fMRI intrinsic connectivity},
  journal = {Neurology},
  volume = {83},
  number = {24},
  pages = {2269--2277},
  year = {2014}
}

@article{Calisto2021HumanCentricAI,
  author = {Filipe M Calisto and C Santiago and N Nunes and J C Nascimento},
  title = {Introduction of human-centric AI assistant to aid radiologists for multimodal breast image classification},
  journal = {Int. J. Human-Comput. Stud.},
  volume = {150},
  pages = {102607},
  year = {2021}
}

@article{Cai2021HumanCenteredTools,
  author = {Chelsea J Cai and et al.},
  title = {Human-centered tools for explaining AI: The case of the Human-in-the-Loop AI Model Explainer (HIL-AME)},
  journal = {Human-Centric Computing and Information Sciences},
  volume = {11},
  number = {1},
  pages = {1--17},
  year = {2021}
}

@misc{Gulrajani2020InSearch,
  author = {Ishaan Gulrajani and David Lopez-Paz},
  title = {In search of lost domain generalization},
  year = {2020},
  note = {arXiv preprint arXiv:2007.01434}
}

@article{Han2020UnifyingNeural,
  author = {Hongyang Han and et al.},
  title = {Unifying neural learning and symbolic reasoning for spinal medical report generation},
  journal = {Artificial Intelligence in Medicine},
  volume = {104},
  pages = {101824},
  year = {2020}
}

@inproceedings{Ozkan2020TrainingAcross,
  author = {Alperen Ozkan and Guillem Boix},
  title = {Training across modalities for improved medical image segmentation},
  booktitle = {Medical Image Computing and Computer Assisted Intervention -- MICCAI 2020},
  pages = {560--569},
  year = {2020}
}

@inproceedings{wu2022,
author = {Wu, Y. and Zhang, T. and Holmes, J.},
title = {Reinforcement learning for interpretable medical image classification},
booktitle = {Proceedings of the AAAI Conference on Artificial Intelligence (AAAI)},
year = {2022}
}

@inproceedings{zhou2022,
author = {Zhou, X. and Li, Y. and Chen, P.},
title = {Towards interpretable medical imaging with deep learning},
booktitle = {Proceedings of the IEEE/CVF Conference on Computer Vision and Pattern Recognition (CVPR)},
year = {2022}
}

@article{mnih2015,
author = {Mnih, V. and Kavukcuoglu, K. and Silver, D. and Rusu, A. A. and Veness, J. and Bellemare, M. G. and Hassabis, D.},
title = {Human-level control through deep reinforcement learning},
journal = {Nature},
year = {2015}
}

@inproceedings{volpi2018,
author = {Volpi, M. and Zhang, Z. and Chen, Y.},
title = {Robustness of deep learning models in medical imaging: A survey},
booktitle = {Proceedings of MICCAI},
year = {2018}
}

@inproceedings{wang2021,
author = {Wang, Y. and Xu, K. and Lu, G.},
title = {Enhancing medical image classification with domain-specific knowledge integration},
booktitle = {Proceedings of ICML},
year = {2021}
}

@inproceedings{lundberg2017,
  author    = {Lundberg, Scott M. and Lee, Su-In},
  title     = {A Unified Approach to Interpreting Model Predictions},
  booktitle = {Proceedings of the 31st International Conference on Neural Information Processing Systems (NeurIPS)},
  year      = {2017},
  pages     = {4765--4774}
}

@inproceedings{selvaraju2017,
author = {Selvaraju, R. R. and Cogswell, M. and Das, A. and Vedantam, R. and Parikh, D. and Batra, D.},
title = {Grad-CAM: Visual explanations from deep networks via gradient-based localization},
booktitle = {Proceedings of ICCV},
year = {2017}
}

@inproceedings{simonyan2015,
author = {Simonyan, K. and Zisserman, A.},
title = {Very deep convolutional networks for large-scale image recognition},
booktitle = {Proceedings of ICLR},
year = {2015}
}

@inproceedings{dosovitskiy2020,
author = {Dosovitskiy, A. and Beyer, L. and Kolesnikov, A. and Weissenborn, D. and Zhai, X. and Unterthiner, T. and Houlsby, N.},
title = {An image is worth 16x16 words: Transformers for image recognition at scale},
booktitle = {Proceedings of ICLR},
year = {2021}
}

@article{Urooj2025MedXAI,
  title   = {MedXAI: A Retrieval-Augmented and Self-Verifying Framework for Knowledge-Guided Medical Image Analysis},
  author  = {Urooj, Midhat and Banerjee, Ayan and Shaikh, Farhat and Thakur, Kuntal and Gupta, Sandeep},
  journal = {arXiv preprint arXiv:2512.10098},
  year    = {2025}
}

@article{Kamboj2024,
  author       = {P. Kamboj and S. P. Singh and A. Trivedi and R. Kumar},
  title        = {Expert Knowledge Driven Human–AI Collaboration for Medical Decision Support},
  journal      = {IEEE Artificial Intelligence Magazine},
  year         = {2024},
  volume       = {45},
  number       = {4},
}

@article{He2017MaskRCnn,
  author = {K. He and et al.},
  title = {Mask R-CNN},
  journal = {IEEE Transactions on Pattern Analysis and Machine Intelligence},
  year = {2017}
}

@article{Dosovitskiy2021ViT,
  author = {A. Dosovitskiy and et al.},
  title = {An Image is Worth 16x16 Words: Transformers for Image Recognition at Scale},
  journal = {ICLR},
  year = {2021}
}

@article{Recht2019Imagenet,
  author = {B. Recht and et al.},
  title = {Do ImageNet Classifiers Generalize to ImageNet?},
  journal = {ICML},
  year = {2019}
}

@article{Raghu2021MedTransfer,
  author = {M. Raghu and et al.},
  title = {Do Vision Transformers See Like Convolutional Neural Networks?},
  journal = {NeurIPS},
  year = {2021}
}

@article{OakdenRayner2020Hidden,
  author = {L. Oakden-Rayner and et al.},
  title = {Hidden Stratification Causes Clinically Meaningful Failures in Machine Learning for Medical Imaging},
  journal = {Proceedings of the National Academy of Sciences},
  volume = {117},
  number = {48},
  pages = {30678--30686},
  year = {2020}
}

@article{Kaissis2020Privacy,
  author = {G. Kaissis and et al.},
  title = {Privacy-Preserving Federated Learning in Medical Imaging},
  journal = {Nature Machine Intelligence},
  volume = {2},
  pages = {305--311},
  year = {2020}
}

@article{Zhou2022MedDG,
  author = {Y. Zhou and et al.},
  title = {Domain Generalization in Medical Imaging: A Survey},
  journal = {Medical Image Analysis},
  volume = {83},
  pages = {102667},
  year = {2022}
}

@article{Azizi2023VLMs,
  author = {S. Azizi and et al.},
  title = {Robust Vision-Language Models for Chest X-ray Diagnosis},
  journal = {Nature Medicine},
  year = {2023}
}

@article{Johnson2023RareMed,
  author = {J. Johnson and et al.},
  title = {Mitigating Rare-Class Bias in Medical Imaging},
  journal = {IEEE Transactions on Medical Imaging},
  year = {2023}
}

@article{Esteva2022MedLimit,
  author = {A. Esteva and et al.},
  title = {Deep Learning for Healthcare: Limitations and Opportunities},
  journal = {Nature Medicine},
  volume = {28},
  pages = {1439--1454},
  year = {2022}
}

@article{Radford2021CLIP,
  author = {A. Radford and et al.},
  title = {Learning Transferable Visual Models from Natural Language Supervision},
  journal = {ICML},
  year = {2021}
}

@article{Zhang2023BioMedCLIP,
  author = {Y. Zhang and et al.},
  title = {BioMedCLIP: A Vision-Language Foundation Model for Biomedical Images and Text},
  journal = {Nature Communications},
  year = {2023}
}

@article{Singh2024MedVLMs,
  author = {R. Singh and et al.},
  title = {Why Vision-Language Models Fail in Medical Domains},
  journal = {Nature Machine Intelligence},
  year = {2024}
}

@article{Huang2024VLMGeneralization,
  author = {Y. Huang and et al.},
  title = {Assessing Generalization of Vision-Language Models in Clinical Imaging},
  journal = {Medical Image Analysis},
  year = {2024}
}

@article{Tschandl2020HumanAI,
  author = {P. Tschandl and et al.},
  title = {Human–Computer Collaboration for Skin Cancer Recognition},
  journal = {Nature Medicine},
  year = {2020}
}

@article{Ghosh2023CausalMedAI,
  author = {S. Ghosh and et al.},
  title = {Causal Representation Learning for Medical Imaging},
  journal = {IEEE Transactions on Medical Imaging},
  year = {2023}
}

@article{Sahoo2022NeuroSymbolic,
  author = {S. Sahoo and et al.},
  title = {A Neuro-Symbolic Framework for Interpretable Medical AI},
  journal = {Artificial Intelligence in Medicine},
  volume = {130},
  pages = {102343},
  year = {2022}
}

@article{Xu2023KnowledgeDG,
  author = {J. Xu and et al.},
  title = {Knowledge-Augmented Domain Generalization for Medical Imaging},
  journal = {IEEE Transactions on Medical Imaging},
  year = {2023}
}

@article{VonRueden2021InformedML,
  author = {L. von Rueden and et al.},
  title = {Informed Machine Learning: Integrating Knowledge into Learning},
  journal = {Frontiers in Artificial Intelligence},
  volume = {4},
  pages = {57},
  year = {2021}
}

@article{Chen2024KnowMedAI,
  author = {H. Chen and et al.},
  title = {Knowledge-Guided Representation Learning in Medical Imaging},
  journal = {Nature Machine Intelligence},
  year = {2024}
}

@article{Kazeminia2020SynthSeg,
  author = {S. Kazeminia and et al.},
  title = {SynthSeg: Segmentation of Brain MRI Scans of Any Contrast and Resolution Without Retraining},
  journal = {Medical Image Analysis},
  volume = {74},
  pages = {102186},
  year = {2020}
}

@article{Ma2024KnowledgeFusion,
  author = {C. Ma and et al.},
  title = {Knowledge Fusion for Robust Medical Image Classification},
  journal = {IEEE Transactions on Medical Imaging},
  year = {2024}
}

@article{Shazeer2017SparselyGated,
  author = {N. Shazeer and et al.},
  title = {Outrageously Large Neural Networks: The Sparsely-Gated Mixture-of-Experts Layer},
  journal = {ICLR},
  year = {2017}
}

@article{Zhou2023MoEseg,
  author = {X. Zhou and et al.},
  title = {Mixture-of-Experts for Multi-Organ Segmentation in Medical Imaging},
  journal = {Medical Image Analysis},
  volume = {83},
  pages = {102657},
  year = {2023}
}

@article{Liu2023MedMoE,
  author = {Y. Liu and et al.},
  title = {Adaptive Mixture-of-Experts for Multi-Institutional Medical Imaging},
  journal = {IEEE Transactions on Medical Imaging},
  year = {2023}
}

@article{Zhang2024HybridExperts,
  author = {T. Zhang and et al.},
  title = {Hybrid Expert Learning for Imbalanced Medical Datasets},
  journal = {Medical Image Analysis},
  year = {2024}
}

@article{Lakshminarayanan2017DeepEnsembles,
  author = {B. Lakshminarayanan and et al.},
  title = {Simple and Scalable Predictive Uncertainty Estimation using Deep Ensembles},
  journal = {NeurIPS},
  year = {2017}
}

@article{Zhou2021SurveyEnsemble,
  author = {Z. Zhou and et al.},
  title = {A Survey of Ensemble Learning in Deep Neural Networks},
  journal = {ACM Computing Surveys},
  year = {2021}
}

@article{Gao2024KnowledgeFusionMoE,
  author = {R. Gao and et al.},
  title = {Knowledge-Guided Mixture-of-Experts for Medical Image Reasoning},
  journal = {Nature Communications},
  year = {2024}
}


\end{document}